\documentclass[11pt]{article}

\usepackage{ulem}

\usepackage[final]{acl}

\usepackage{times}
\usepackage{latexsym}
\usepackage{amssymb}
\usepackage{paralist}
\usepackage{todonotes}

\usepackage[T1]{fontenc }

\usepackage[utf8]{inputenc}

\usepackage{microtype}

\usepackage{inconsolata}

\usepackage{graphicx}
\usepackage{booktabs}
\usepackage[table]{xcolor}
\usepackage{tabularx}
\usepackage{array}
\usepackage{multirow}
\usepackage{makecell}
\usepackage{helvet}
\usepackage{parskip}
\usepackage{caption}
\usepackage{longtable}
\usepackage{hyperref}
\usepackage[most]{tcolorbox}
\usepackage{makecell}
\usepackage{todonotes}

\definecolor{notrecommended}{HTML}{FF6961}
\definecolor{conditional}{HTML}{F8D66D}
\definecolor{acceptable}{HTML}{7ABD7E}
\definecolor{secScreen}{HTML}{2C3E50}   
\definecolor{secExp}{HTML}{1A5276}      
\definecolor{secOpin}{HTML}{1E8449}     
\definecolor{secDemo}{HTML}{6E2F8B}     
\definecolor{secOther}{HTML}{7B5D3A}    
\definecolor{rowScreen}{HTML}{e7eef4}
\definecolor{rowExp}{HTML}{D6E4F0}
\definecolor{rowOpin}{HTML}{D5F5E3}
\definecolor{rowDemo}{HTML}{E8DAEF}
\definecolor{rowOther}{HTML}{F5EEE6}
\definecolor{condCol}{HTML}{922B21}     

\usepackage{longtable}
\usepackage{amssymb}
\renewcommand{\paragraph}[1]{\par\noindent\textbf{#1}}
\usepackage{xspace}
\usepackage{wasysym}
\usepackage{pifont}

\definecolor{pastelgreen}{RGB}{223,239,216}
\definecolor{pastelorange}{RGB}{245,229,204}
\definecolor{pastelred}{RGB}{242,214,214}

\newcommand{\cond}[1]{\textcolor{condCol}{\textit{\footnotesize #1}}}
\newcommand{\sechead}[2]{%
  \multicolumn{4}{l}{%
    \cellcolor{#1}\textbf{\textcolor{white}{\large #2}}%
  }\\[-2pt]
}

\newcommand\blfootnote[1]{%
  \begingroup
  \renewcommand\thefootnote{}\footnote{#1}%
  \addtocounter{footnote}{-1}%
  \endgroup
}
\title{Can Crowdsourcing Survive the LLM Era?\\ A Community Survey on Human Data Collection}

\author{
Aswathy Velutharambath\textsuperscript{\ding{168}\ding{115}}, 
Neele Falk\textsuperscript{\ding{168}}$^{\dagger}$, 
Sofie Labat\textsuperscript{\ding{169}\ding{170}}$^{\dagger}$, \\
\textbf{Tarun Tater}\textsuperscript{\ding{168}}$^{\dagger}$ \and
\textbf{Amelie W{\"u}hrl}\textsuperscript{\ding{171}}$^{\dagger}$ \\
\textsuperscript{\ding{115}}KU Leuven, Belgium
\textsuperscript{\ding{168}}University of Stuttgart, Germany, 
\textsuperscript{\ding{169}}Ghent University, Belgium \\
\textsuperscript{\ding{170}}Harvard University, USA,
\textsuperscript{\ding{171}}IT University of Copenhagen, Denmark\\
\small\texttt{Corresponding author: aswathy.velutharambath@kuleuven.be}\\
}

\begin{document}
\maketitle
\blfootnote{$^{\dagger}$Equal contribution.}
\begin{abstract}

The widespread use of large language models (LLMs) as writing tools challenges the validity of crowdsourced data, as crowdworkers may `outsource' tasks to models.
To better understand how the research community addresses this, we surveyed 155 researchers in NLP and related disciplines about their experiences and opinions on collecting free-text responses via crowdsourcing in the LLM era. This paper provides an overview of practitioners' challenges, mitigation strategies, and the foreseen implications on data quality. 
44\% of respondents reported observing LLM usage in their
crowdsourced data. While 93.2\% of them had anticipated this, half were
unsure what precautions to take. The most prevalent detection
strategies are distinctive textual style patterns and unusually fast completion
times. Overall, survey responses show that the research community is aware of the problem and taking measures, but existing efforts
remain insufficient to fully address it. 
Finally, we derive a set of considerations 
to guide future crowdsourced free-text data collection in the era of LLMs.
We make the survey and an anonymized version of the responses available for researchers\footnote{Dataset is available from: \url{https://doi.org/10.7802/3102}.}.

\end{abstract}

\section{Introduction}

In crowdsourcing studies, researchers collect diverse data from a large group of individuals, usually via online platforms. The data collection method is widely used across a variety of fields such as natural language processing (NLP), computational social science, psychology, and related areas \cite{munro-etal-2010-crowdsourcing, Bigham2017Crowd, wazny2017crowdsourcing, suhr-etal-2021-crowdsourcing}. 
In NLP, crowdsourced data is used to benchmark a range of tasks such as sentiment analysis, emotion detection, natural language inference, and question answering \cite{socher-etal-2013-recursive, bowman-etal-2015-large, rajpurkar-etal-2016-squad, suhr-etal-2021-crowdsourcing, troiano-etal-2024-dealing}. The approach assumes that workers contribute diverse, independent human inputs that reflect a variety of perspectives, language styles, and reasoning strategies. However, as large language models (LLMs) become widely used everyday tools, crowdsourcing faces critical challenges. Crowdworkers can now easily access generative tools to quickly produce fluent, high-quality text for almost any task. 

\begin{figure}
    \centering
    \includegraphics[width=1\linewidth]{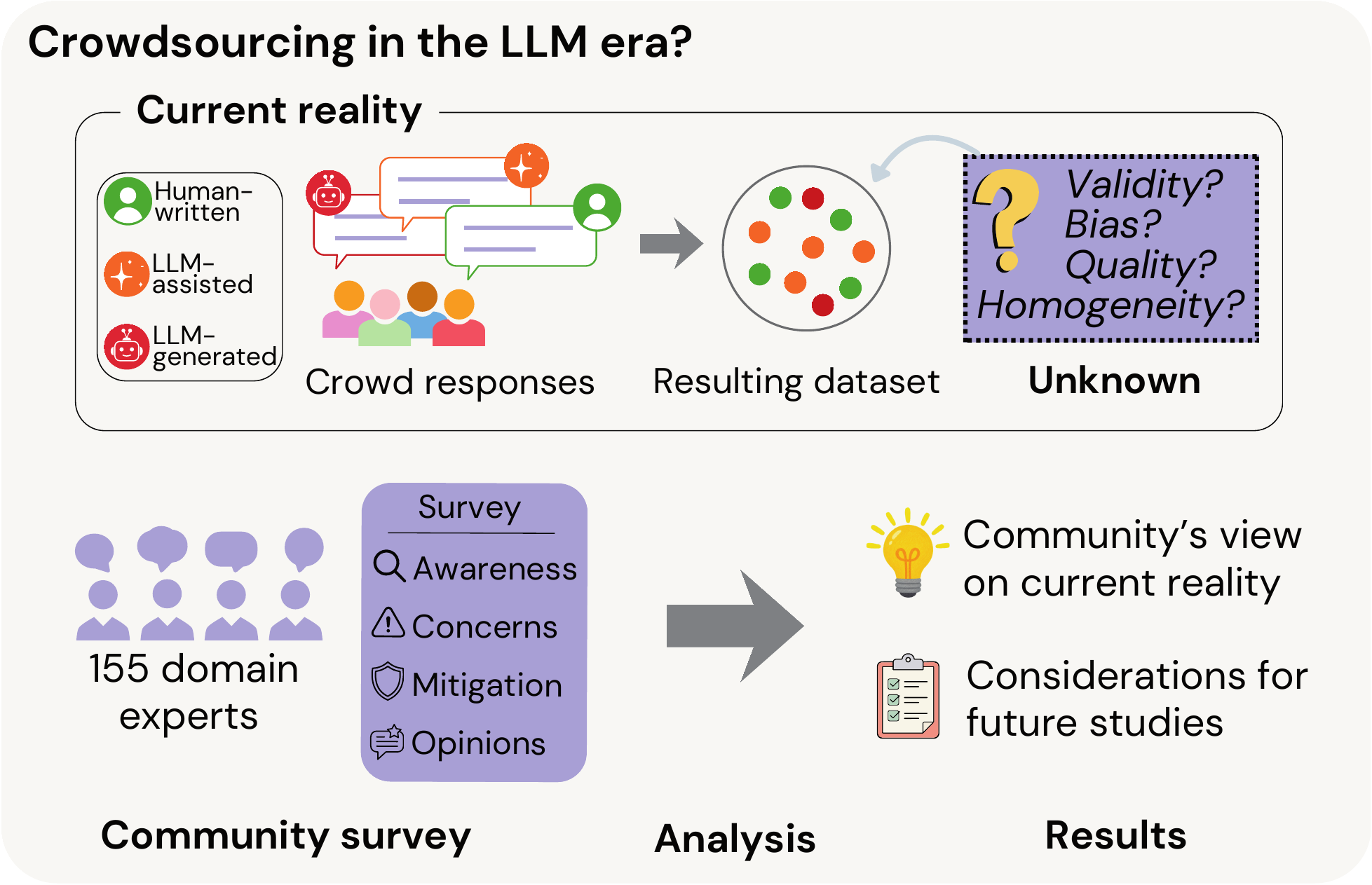}
    \caption{Survey on LLM use in crowdsourced data.}
    \label{fig:placeholder}
\end{figure}

Undisclosed LLM use by crowdworkers can undermine the quality and interpretability of crowdsourced data \cite{veselovsky2025prevalence, zhang2025generative}, posing risks for studies in NLP, human--computer interaction, and social sciences that rely on authentic human perspectives, behaviors, and language use. 
For downstream applications, LLM-generated text tends to exhibit reduced diversity and more homogeneous patterns, potentially reinforcing existing biases and contributing to model collapse when used to train systems \cite{kamruzzaman-2024-investigating,wright2026epistemicdiversityknowledgecollapse}. To discourage LLM use, researchers have adopted ad-hoc mitigation  
strategies such as including explicit instructions, adding technical hurdles, and resorting to
in-house annotation \citep{veselovsky2023ai, kieslich2025anticipating,
Zhong23}. 

As LLMs become more prevalent, AI-assisted content is likely to appear not only in crowdsourced studies but also in naturally occurring web data, limiting the effectiveness of simply relying on ``real-world'' sources. Accordingly, whether LLM usage should be considered problematic depends on the study's goals: while it may undermine efforts to capture authentic human judgments or behavior, it may pose fewer concerns in factual, retrieval-based, or explicitly human--AI collaborative tasks. Despite the growing prevalence of these concerns, we still lack a systematic understanding of how researchers in NLP and adjacent disciplines experience and respond to LLM usage in crowdsourcing.

To address this gap, we survey researchers who have worked with crowdsourced  
data to examine their experiences and views on LLM usage by crowdworkers, specifically in the context of free-text responses. We focus on three questions: (RQ1)~To what extent are LLM-generated responses a concern in 
crowdsourcing studies, and how do researchers address them?
(RQ2)~How do researchers perceive the impact of LLM-generated answers in crowdstudies?
(RQ3)~Which practical considerations emerge from the community's experiences for running crowdstudies in the LLM era?

To the best of our knowledge, this is the first systematic community
survey on LLM use by crowdworkers in free-text responses.. We consolidate practitioners' experiences
and perspectives to establish a shared understanding of
a problem 
previously only addressed through individual,
ad-hoc practices. Finally, we derive actionable considerations to guide future 
crowdsourcing studies.

\section{Background and Related Work}
\subsection{Crowdsourcing: Paradigm \& Challenges}
Using non-experts to annotate data is not new:
\citet{vonahn2004labeling} demonstrated that web users produce reliable
labels via games, and \citet{snow2008cheap}
 showed that aggregated non-expert judgments can match expert quality across NLP tasks, driving widespread adoption in building training corpora and benchmarks 
\citep[i.a.,][]{callison-burch-dredze-2010-creating,
bowman-etal-2015-large, rajpurkar-etal-2016-squad}. 
Since then, a substantial body of work has examined annotation quality, emphasizing that task setup \cite{bhardwaj-etal-2010-anveshan}, annotation aggregation \citep{Sheng-etal-2009, hsueh-etal-2009-data}, worker selection \cite{roit-etal-2020-controlled}, and quality control \cite{roit-etal-2020-controlled, klie-etal-2024-analyzing} can critically affect the creation of reliable datasets via crowdsourcing.

Crowdsourcing platforms are also used 
in behavioral and social science research to collect free-text responses,
personality measures, and survey data at scale
\citep{buhrmesterAmazonsMechanicalTurk2011,
millerUsingOnlineCrowdsourcing2017, peer2017beyond}. Regardless, prior work highlights methodological, ethical, and
practical challenges in task design, execution, and
evaluation, as well as broader concerns for crowdsourcing in NLP
\citep{nouri-etal-2020-mining, Yunhui2021, shmueli-etal-2021-beyond}.

The widespread availability of LLMs introduces new challenges for studies relying on free-text responses.
\citet{veselovsky2023ai} estimate that 30--40\% of crowdworkers rely
on LLMs for text production, and \citet{zhang2025generative}
report that 34\% of Prolific participants self-report using LLMs for
open-ended questions. Concerns about undisclosed LLM usage are
increasingly recognized as a limitation in crowdsourced studies
\citep[i.a.,][]{moller-etal-2024-parrot, trott2024can,
gligoric-etal-2025-unconfident}.
Mitigation strategies such as disabling copy-paste, presenting prompts
as images, and explicit instructions have been proposed
\citep{kieslich2025anticipating}, but can reduce
rather than eliminate LLM-assisted responses
\citep{veselovsky2025prevalence}. Some studies avoid crowdsourced
evaluation 
in favor of expert or author-based assessment
\citep{Zhong23}.

\subsection{Human vs.\ LLM Language and Behavior} 
LLM outputs
exhibit lower lexical diversity \citep{zanotto24, Akinwande_2024,
YildizDurak2025, Culda31122025}, more positive sentiment \citep{MuozOrtiz2024,
zamaraeva-etal-2025-comparing}, and distinctive lexical and stylistic
patterns compared to human-written language \citep{Gray2024ChatGPTE, Kobak2025, Reinhart2025}. LLM
assistance increases similarity across authors and reduces lexical and
content diversity \citep{Padmakumar2024}, including in narrative tasks
\citep{begus2024experimental} and survey settings
\citep{veselovsky2025prevalence, zhang2025generative}. LLMs can also
alter the underlying meaning and voice of text even with light editing
\citep{abdulhai2026llmsdistortwrittenlanguage}, suppress underrepresented
perspectives \citep{Burton2024}, and risk model collapse and reduced
epistemic diversity in downstream applications
\citep{Shumailov2024_AI, wright2026epistemicdiversityknowledgecollapse}.
LLMs further encode systematic biases and remain prone to hallucinations
\citep{kamruzzaman-2024-investigating, Ji2023}, threatening the validity
of any downstream analysis that treats these responses as authentic
human-generated text.

\subsection{Automatic Detection of LLM Usage}

While LLM and human
language differ systematically, reliable detection remains difficult. LLM-generated responses can pass standard quality checks \cite{westwood2025}. Existing detection methods, such as watermarking \citep{watermarking},
transformer-based classifiers \citep{Guggilla2025AIGT}, and
model-related or linguistic features \citep{wu-etal-2025-survey,
hamedDetectionChatGPTFake2024}, degrade
substantially under paraphrasing attacks and in out-of-domain settings
\citep{XIANG2026, wu-etal-2025-survey}. 
Detecting human--AI co-authored text is harder still, as models trained
on fully human or AI-generated text transfer poorly to
mixed-authorship settings \citep{richburg-etal-2024-automatic,
su-etal-2025-haco}. While datasets such as CoAuthor \citep{coauthor} and
more recent benchmarks \citep{coauthorbenchmark} begin to address
this, fine-grained co-authored text detection remains far from
solved \citep{su-etal-2025-haco}. In sum, reliably distinguishing authentic
human responses from LLM-assisted ones in crowdsourced free-text
settings is not yet feasible.

\section{Survey Methodology}
\label{sec:survey_method}
We used Qualtrics\footnote{https://www.qualtrics.com} to design an online survey, which was conducted from 15 August 2025 to 15 March 2026. 
We distributed the survey via social media (e.g., LinkedIn, Bluesky), mailing lists (e.g., ACL portal, Corpora List), personal contacts, and cold emails to authors of recent *ACL publications related to crowdsourcing (see Appendix~\ref{append:contact authors}). As the ACL Anthology was a primary recruitment source, our findings are primarily representative of the NLP community, although respondents from adjacent disciplines are also included.

While our focus is on researchers with first-hand crowdsourcing experience, we also gather perspectives from those without. Given the prevalence of LLM use in general, we presume community members have formed opinions that are valuable to investigate, even without first-hand crowdsourcing experience. The survey is structured into three sections: experiences (for participants with prior crowdsourcing
experience only), opinions, and respondent information. Questions were presented in single-choice, multiple-choice,
or free-text formats (see Appendix \ref{append:survey-flow} for questions and workflow).

\paragraph{Experiences.}
Shown only to respondents with prior free-text crowdsourcing experience, this section
covers background information (experience, platforms,
languages, research goals), use of free-text data, and analyses conducted. It also includes questions on
experiences with crowdworker LLM usage: observed cases, indicators, handling of affected data/workers, precautions taken, and perceived impact on research.

\paragraph{Opinions.} Respondents share their views on LLM-generated text in research, potential methods to discourage LLM usage, expectations from crowdsourcing platforms, acceptable use conditions for LLM usage, and any additional thoughts.

\paragraph{Respondents' background.} We collect information on respondents' demographic characteristics and their professional background, including scientific discipline\footnote{The categories for scientific disciplines follow \href{https://www.dfg.de/en/research-funding/proposal-funding-process/interdisciplinarity/subject-area-structure}{the German Research Foundation (DFG) classification.}} and research role.

\section{Data Analysis}

\paragraph{Overview.} In total, $401$ people accessed the survey, of whom $162$ completed it. After excluding responses with a completion time of less than two minutes and one non-consenting case, a total of $155$ responses remained for analysis.
The median completion time was $13.8$ minutes. Nine respondents ($5.8\%$) self-reported using generative AI to fill parts of the survey, highlighting the pervasiveness of the phenomenon even in meta-research contexts.\footnote{We do not explicitly instruct participants to refrain from using LLMs.} 

\subsection{Information on Respondents}
We first characterize our sample, as the distribution of respondents across demographics, research profiles, and prior crowdsourcing experience provides the necessary context to interpret the findings reported in the following sections.

\begin{figure}[h!]
 	\center
     \includegraphics[width=1\linewidth]{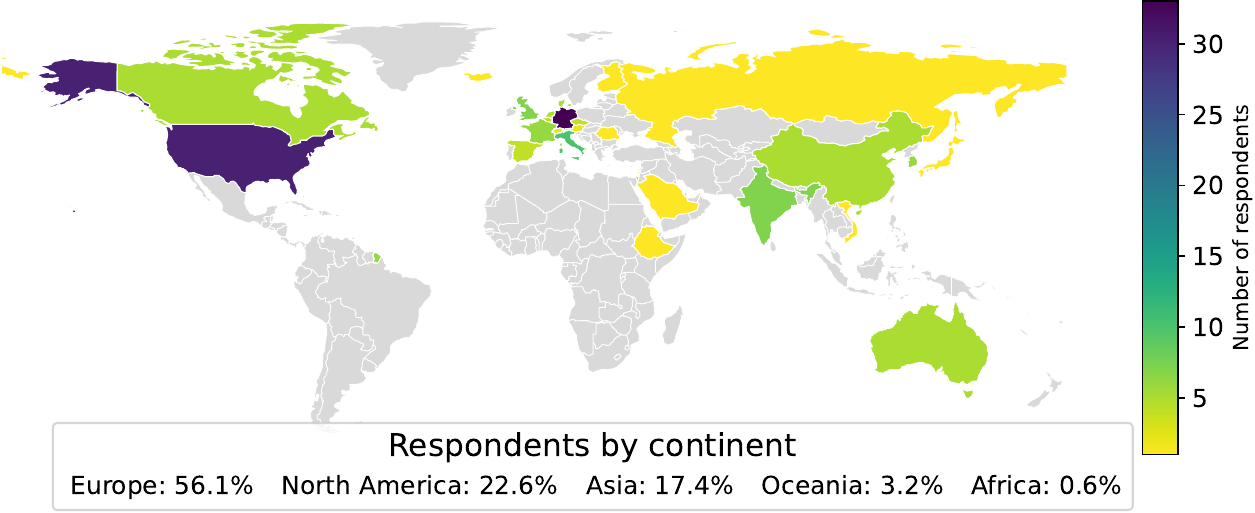}
    \caption{Geographic distribution of respondents based on country of employment.}
 	\label{fig:country}
 \end{figure}

\paragraph{Demographics.}
Among the respondents, 53.5\% identify as men ($n=83$), 37.4\% as women ($n=58$), and 2.0\% as non-binary ($n=3$). 7.1\% ($n=11$) preferred not to disclose their gender. Furthermore, the respondent pool is strongly skewed toward the 25--29 and 30--34 age groups (60.6\%, $n=94$), accounting for the majority of participants. Representation declines steadily across older age categories. As for geographic distribution based on country of employment (see Fig.~\ref{fig:country}), respondents are concentrated in Europe (56.1\%, $n=87$), with North America (22.6\%, n $=35$) and Asia (17.4\%, $n=25$) contributing the next largest share. At the country level, Germany (21.3\%, $n=33$) and the United States (19.4\%, $n=30$) dominate, while a tail of lower participation from other countries reflects a broad, but uneven international reach.

\begin{figure}[!h]
 	\center  \includegraphics[width=\columnwidth]{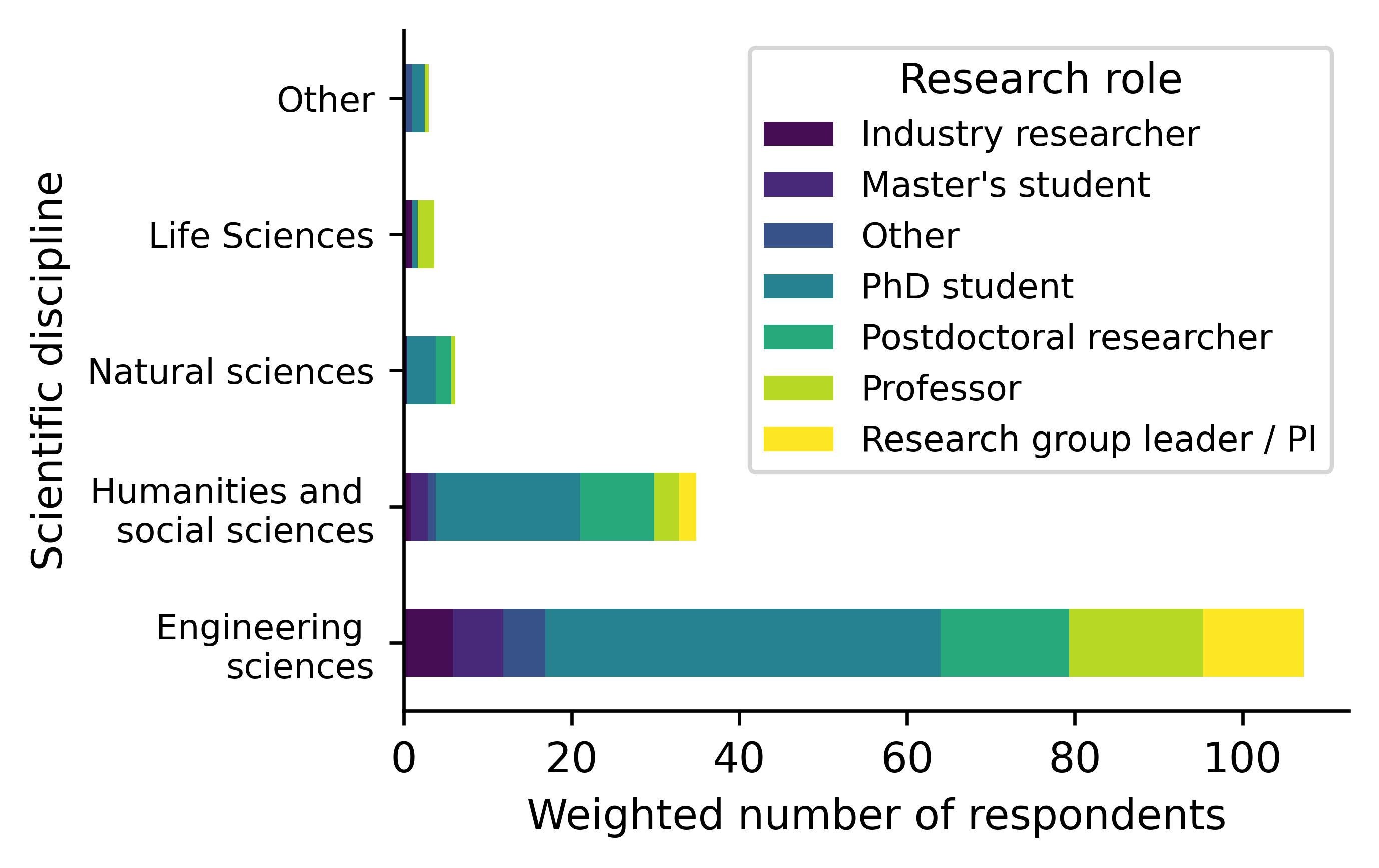}
    \caption{Respondents per scientific discipline and research role, weighted for multiple selections.}
  
 	\label{fig:research_area}
 \end{figure}

\paragraph{Research profiles.} 
Fig.~\ref{fig:research_area} illustrates that our sample is dominated by respondents from the engineering sciences (69.3\%), with humanities and social sciences forming the second largest group (22.5\%), and only limited representation from the natural (4.0\%) and life sciences (2.4\%). The skew toward engineering sciences is expected given our *ACL-based recruitment strategy. Across disciplines, respondents are mainly early-career researchers, especially PhD students, while postdocs, professors, and PIs constitute a substantial second group.

\paragraph{Crowdsourcing experience.} Among the $155$ respondents, $100$ ($64.5\%$) had experience conducting free-text crowdsourcing studies. Between them, the most common study goals were evaluation of model outputs (58.0\%), annotation or coding for qualitative analysis (45.0\%), training machine learning models (38.0\%), data validation or quality assurance (36.0\%), behavioral or cognitive studies (35.0\%), and opinion surveys (33.0\%). The most commonly used platforms were Prolific (53.0\%) and Amazon Mechanical Turk (MTurk) (47.0\%), and respondents represented all levels of crowdsourcing experience, from less than one year to more than five years.

\subsection{Prevalence, Risks, and Mitigation of LLM-Generated Responses (RQ1)} \label{sec: RQ1}

\begin{figure*}
    
    \centering
    \includegraphics[width=1\linewidth]{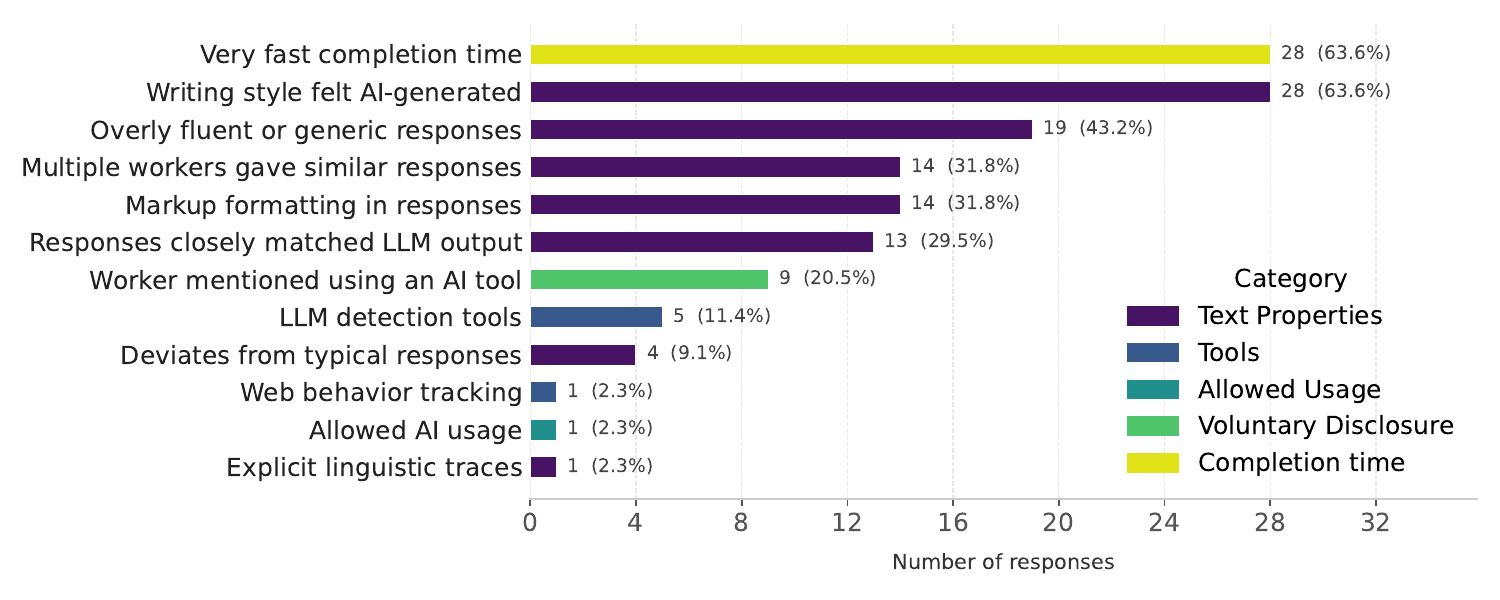}
    \caption{Signals of LLM usage grouped into categories.}
    \label{fig:signals}
\end{figure*}

Of the $100$ respondents with prior free-text crowdsourcing experience, only 44 reported observing LLM-generated responses. This finding may reflect either effective study designs that discouraged LLM use or cases where such use went undetected or was not actively assessed. Supporting the latter
interpretation, observation rates increased with experience:
researchers with over five years of crowdsourcing experience
reported higher rates of observed LLM use ($60.0$\%) than those with one to two years ($\chi^2 = 17.64$, $p = 0.007$,
$df = 6$). 

In contrast, disciplinary background showed no significant association with reported observations of LLM use (Fisher's exact test for each discipline): Engineering sciences ($p = 0.963$), Natural sciences ($p = 0.431$), Humanities and Social sciences ($p = 0.186$), Life Sciences ($p = 1.000$), and Other ($p = 0.645$).\footnote{Because disciplines were multiple-choice, these five tests are not fully independent.}

\paragraph{Signals of LLM usage.}
As shown in Fig.~\ref{fig:signals}, among the $44$ researchers who observed LLM usage, the most common indicators were textual properties: AI-like writing style ($63.6\%$), overly fluent or generic responses ($43.2\%$), high similarity across multiple workers ($31.8\%$), markup formatting ($31.8\%$), and close resemblance to known LLM output ($29.5\%$). Among these, markup formatting is a relatively clear signal, 
while other textual signals involve more subjective judgments. Participants also note that careful prompt engineering can make LLM output closely resemble human writing, making subtle cases difficult to detect.

$11.4\%$ ($n=5$) of researchers used automated LLM detection tools, though no universally reliable detection system currently exists. Without knowing the model, flagged responses may be indistinguishable from human writing, making decisions based on them as unreliable as subjective judgments. Unusually fast task completion was mentioned by $63.6\%$ ($n=28$) as a signal. Regardless of whether LLM usage is the underlying cause, this serves as a robust indicator of low-effort responses that warrant exclusion. Finally, some researchers relied on more direct forms of evidence: $20.5\%$ ($n=9$) reported workers voluntarily disclosing their use of an AI tool, and $2.3\%$ ($n=1$) explicitly allowed AI usage. Both voluntary disclosure and explicit allowance reflect transparent use, offering the clearest basis for action.

\paragraph{Precautions taken during study design.}
Among the $44$ who observed LLM usage, most ($93.2\%$) anticipated this possibility prior to data collection, and $75.6\%$ of them took precautions in their study design. 
Common measures included warnings about LLM usage ($67.7\%$) and attention checks ($61.3\%$). While intuitive, the effectiveness of warnings is difficult to assess \cite{bruhlmannEffectivenessWarningStatements2024}. Moreover, warnings may have unintended effects by making participants aware of or even prompting the very behavior they aim to discourage. More restrictive measures, such as disabling copy-paste ($32.3\%$) and adding CAPTCHA tasks ($19.4\%$), raise the effort required to use LLMs, but their effectiveness still hinges on whether the perceived benefit of doing so outweighs the inconvenience of bypassing these restrictions. Among the $10$ researchers who took no precautions, half were unsure how to address the issue, indicating a knowledge gap in practical and accessible solutions.

\paragraph{Handling detected LLM usage.}
Upon detecting LLM-generated responses, researchers describe two strategies. 
On the data side, $56.8\%$ excluded such instances, $27.3\%$ included them with a label flagging it as potentially LLM-generated, and $25.0\%$ set them aside for separate analysis. Another $13.6\%$ included them without special treatment, raising concerns about LLM-generated content silently entering research pipelines. 
They also describe taking actions against workers. $38.6\%$ rejected submissions and an equal proportion blocked the workers. Others flagged or reported the worker on the platform, withheld payment, or confronted them ($15.9\%$, $n=7$ each). Notably, $29.5\%$ took no action. Beyond the ethical implications of penalizing workers based on unreliable signals and losing valid data, these responses impose a considerable overhead in time and effort on researchers.

These findings show that LLM-generated responses are a real concern in studies which rely on crowdsourcing for free-text data. Yet current detection signals and safeguards are often unverifiable, subjective, or easy to circumvent. This points to a troubling gap between awareness of the problem and the availability of systematic solutions.

\subsection{Perceived Impact of LLM-generated Responses on Data Validity (RQ2)} \label{sec: RQ2}

We ask participants to describe how they perceive the impact of LLM-generated text in crowdsourcing studies on research that relies on human-written data, for instance regarding data quality, validity, impacts on study outcomes.
Rather than focusing only on operational concerns such as detection or prevention, these responses provide broader insights into how crowdsourced free-text data is interpreted in the LLM era.
We conduct a two-stage analysis on the $127$ responses\footnote{We exclude empty responses and responses from participants who self-reported using generative AI during the survey.} we received.  
First, two authors independently inspect the free-text responses using inductive coding to identify recurring themes. In parallel, we prompt \texttt{qwen3.5\_27b}~\cite{qwen3.5} to perform the same theme extraction task (see Appendix~\ref{app:prompts} for the prompt).\footnote{We choose this model as it is a high-performant, open-weight model, feasible to run on our compute infrastructure.} 
We manually aggregate and refine the themes identified by both authors into a shared taxonomy. This results in $26$ themes. 
We then manually review the LLM-extracted themes for additional themes not captured by the manual coding.
The model output surfaced two themes distinct enough from our manual taxonomy to warrant inclusion: \textit{Risk of distorted results} and \textit{Loss of creativity}. Adding these results in a final taxonomy of $28$ themes. Otherwise, the two sets of themes overlapped substantially. Notably though, the raw model output was highly fragmented, producing many near-duplicates for the same underlying theme.
In a second step, we quantify how frequently participants mention these themes across responses. To this end, one author labels each response with one or more theme tags drawn from the consolidated taxonomy.

We group the resulting themes into four thematic groups: \textit{changes in the collected data}, \textit{downstream consequences}, \textit{reactions \& strategies when dealing with LLM-generated content}, and \textit{the acceptability and prevalence of LLM use being task dependent}.
In the following, we describe each group together with the corresponding theme frequencies. 
Figure~\ref{fig:impact_themes} in Appendix~\ref{appendix-rq2} provides a detailed overview of theme frequencies. Table~\ref{tab:themes-impact}, also in Appendix~\ref{appendix-rq2}, complements this by listing all themes together with representative respondent quotes. 

In the first category, respondents describe \textit{changes in the collected data} they observe or anticipate. \textbf{$44.9\%$ ($n = 57$)} of responses convey this major theme (note that responses may express multiple themes within a category).
Respondents frequently characterize LLM-generated responses as more homogeneous ($n = 24$) and less reflective of diverse human perspectives ($n = 16$). Several participants further express concerns that LLM-generated content may amplify existing model biases ($n = 12$), or add noise to the data ($n = 16$). Some respondents also note that LLM assistance can improve stylistic quality, particularly fluency and grammatical consistency ($n = 6$), while others remark that crowdsourced data is generally of low quality, therefore LLM involvement may not substantially worsen the dataset ($n = 2$).
Collectively, these responses predominantly reflect concerns about the potential erosion of human signal and variation in crowdsourced data.

Many respondents also discussed the \textit{downstream consequences} of LLM-generated responses in crowdsourcing studies, which is the most prevalent, appearing in \textbf{$53.5\%$ ($n = 68$)} of responses.
It includes the downstream risk of training future systems on this synthetic data ($n = 12$), the convergence of AI and human data in the future ($n = 3$) and the increased burden on researchers ($3.9\%$, $n = 5$),.
Most prominently though, respondents highlight concerns about result validity ($n = 29$), result reliability ($n = 16$), and distorted study outcomes ($n = 10$).

One third of survey responses discuss \textit{reactions \& strategies when dealing with LLM-generated content} ($n = 43$). Respondents highlight the difficulty of detecting LLM use ($n = 11$), a need to adapt the way we think about data generation ($n = 9$), and the need for guardrails ($n = 8$). They further discuss the need for alternative data collection strategies ($n = 6$), a possible shift in the human role toward verification ($n = 4$), to what extent crowd workers `cheating' is a new development ($n = 4$), and whether LLM use is acceptable when disclosed ($n = 4$).

Finally, respondents discuss how \textit{the acceptability and prevalence of LLM use is task-dependent} ($11.8\%$, $n = 15$). 
Here, the most common position is that the impact of LLM use depends on the goal of the study ($n = 7$). A smaller subset reflects on how LLM use can be acceptable for refinement tasks ($n = 3$). 
They also point out that LLM use is less prevalent when incentives are set consciously, for instance through fair payments, meaningful tasks or tasks that can be actually accomplished in the given time ($n = 2$). Finally, some respondents point out that the problem is particularly harmful for low-resource languages ($n = 3$).

Taken together, these responses suggest that researchers view LLM-generated crowd responses not only as a quality-control issue. Instead, they reflect on it as a broader challenge affecting in which settings the methods can (still) be used and how the data is to be interpreted and validated. 

\subsection{Considerations for Future Studies (RQ3)}\label{rq3}
\begin{figure}
    
    \centering
    \includegraphics[height=6cm,width=\linewidth]{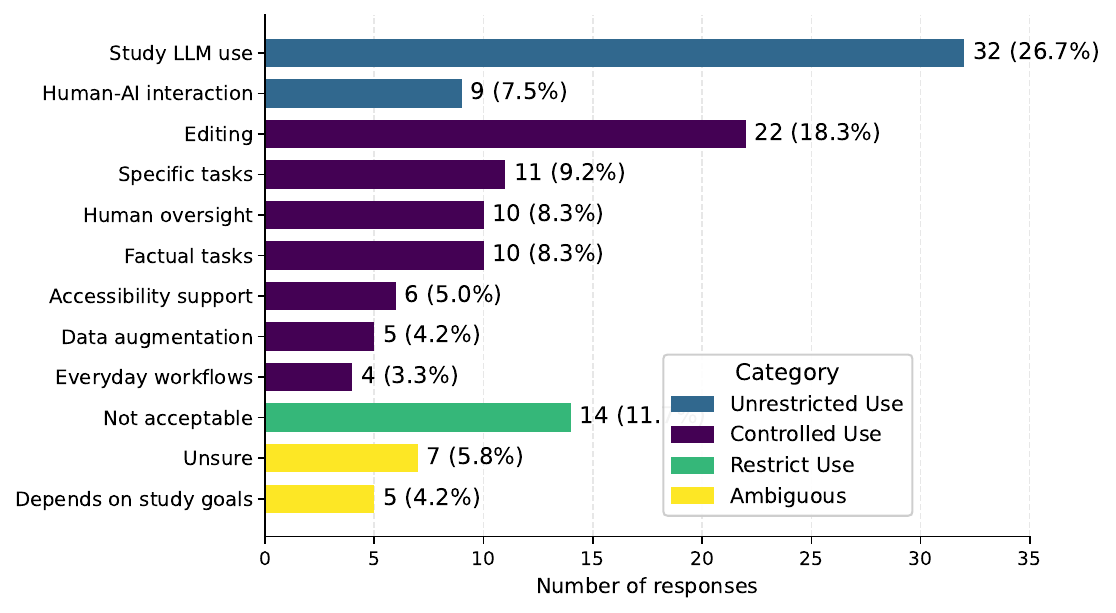}
    \caption{Acceptable LLM usage grouped into categories.}
    \label{fig:acceptable}
\end{figure}

The concerns raised by respondents of our survey are serious and unfortunately not fully resolvable with prescribed, fixed solutions. We therefore ask: what should researchers consider when collecting crowdsourced free-text data in the era of LLM? Based on the experiences and opinions shared in our survey, we organize these
considerations around three stances on LLM acceptability:
fully restricting use, allowing controlled use, and allowing unrestricted use. For each, we discuss implications for study design and post-collection measures, supported where applicable by direct quotes from respondents. Figure~\ref{fig:acceptable} shows the full distribution of acceptability themes from the survey. In addition, Table~\ref{tab:acceptable_themes} in Appendix~\ref{app:acceptable_quotes} provides example quotes from the respondents.

\subsubsection{Restricted LLM Usage}
This is advisable if the study aims to measure attitudes, behavior, personality traits, or personal experience, or when interpersonal, demographic, or cultural variance is meaningful to the research question. Here, as noted by the respondents, LLM use risks ``\textit{masking genuine human perspectives}'' and the resulting ``\textit{data no longer reflects genuine, actual occurring human communication.}'' We observe that $11.7\%$ of respondents ($n=14$) out of $120$ explicitly stated that LLM use is never acceptable in crowdsourcing, often arguing along the lines of \textit{``if LLM-generated output suffices, there is no reason to crowdsource in the first place.''} 

\textbf{Study design considerations.} Studies requiring strict restriction of LLM use face the greatest
methodological challenges, as crowdworkers have broad access to LLM
tools and may use them regardless of task instructions. Warning crowdworkers to 
not use LLMs is necessary (``\textit{stating that the output will be checked for LLM-generated answers and the annotators will somehow not be rewarded that much might help too.}'') but this is insufficient; study design
has to account for likely non-compliance rather than assume
that warnings alone will deter use. 

Where feasible, researchers should actively employ technical measures and hurdles, 
such as disabling copy-paste, presenting stimuli as images and adding CAPTCHA, 
to increase the effort required for undisclosed LLM use. However, such measures should be implemented 
with the understanding that no technical hurdle is entirely insurmountable. 
In addition to technical mitigation strategies, many respondents also emphasized the importance 
of designing tasks that are short, meaningful, and engaging, thereby increasing participants' 
intrinsic motivation to complete the task without relying on LLM assistance.

When authentic human signal is very crucial for the study and prevention mechanisms are deemed insufficient, researchers should evaluate whether more controlled settings, such as laboratory studies or supervised online experiments, are a feasible alternative. As one participant puts it: ``To\textit{ mitigate these issues, I prefer working with a smaller in-house annotation group and regularly reviewing their responses through random checks.}''

Another important consideration is the choice of crowdsourcing platform. While no systematic comparison of
LLM usage rates across platforms exists, the extent to which a platform acknowledges and addresses LLM use may become a useful selection criterion.
Prolific, for instance, allows researchers to enable LLM authenticity checks on free-text
questions.\footnote{\url{https://researcher-help.prolific.com/en/articles/445144-what-are-authenticity-checks}}

Device choice also warrants consideration. Desktop environments make LLM use easier through multitasking and copy-paste, whereas mobile devices create more friction. Restricting participation to mobile may therefore reduce undisclosed LLM use, though the growing integration of AI assistants into smartphones may diminish this advantage over time. Finally, researchers
should be aware that the threat extends beyond assisted use: fully
autonomous agents can complete surveys without any human involvement
and are substantially harder to detect \citep{westwood2025}.

\paragraph{Post-collection considerations.}
Despite precautionary measures, the possibility that some participants may still have used LLMs during task completion cannot be excluded. 
As discussed in Section \ref{sec: RQ1}, researchers commonly rely on indicators such as textual patterns along with short completion times 
to identify potentially LLM-assisted responses. In practice, assessments based on textual patterns are often performed manually and therefore 
remain inherently subjective. While automated detection tools may also be used, current methods do not provide sufficiently reliable 
identification, making some degree of uncertainty unavoidable.

Therefore, any post-collection filtering should be conducted cautiously and reported transparently. 
Researchers should clearly disclose the exclusion criteria and filtering procedures used. Since imperfect filtering may both exclude 
valid responses and retain potentially contaminated instances, sensitivity analyses comparing results before and after filtering 
can help assess the robustness of findings. Ultimately, the best-case scenario is to define and implement effective LLM-use mitigation 
strategies during data collection so that the need for post-collection filtering is minimal.

\subsubsection{Controlled LLM Usage}
Controlled use of LLM could be allowed when the task targets factual or retrievable information rather than subjective opinion or personal experience. In such cases, LLM assistance may not fundamentally compromise the signal of interest. As one survey participants adds ``\textit{it might ease cognitive load of workers, allowing for more or more thorough checking if facilitated well}.'' The largest share of our respondents ($50.8\%, n=61$) identified situations where LLM assistance may be partially acceptable. 
The most frequently cited use cases are editing or grammar polishing ($18.3\%, n=22$): ``\textit{Polishing (in terms of language) user answers that contain their private opinions},'' miscellaneous task-specific contexts such as summarisation or web search ($9.2\%, n=11$), factual or non-subjective tasks ($8.3\%, n=10$), and human oversight workflows where annotators verify LLM-generated drafts ($8.3\%, n=10$). Accessibility support for non-native speakers ($5.0\%, n=6$) and data augmentation in data-scarce domains ($4.2\%, n=5$) were also raised. 

\paragraph{Study design considerations.} When controlled LLM usage is allowed, this should be clearly communicated in the task instructions, including concrete examples of allowed and disallowed uses of LLMs. Furthermore, as noted by one respondent, LLM use should ideally be subject to explicit disclosure requirements, ``\textit{including asking participants to candidly report the method used for producing text for a study},'' so that such usage can be accounted for during analysis. Since participants may rely on different LLM models and to varying degrees, researchers should consider documenting factors such as the tool used, the nature of assistance received (e.g., grammar correction, summarization, drafting), and the extent of reliance on LLM-generated content.

One challenge in this setup is that strict mitigation strategies may no longer be appropriate, since LLM use is intentionally permitted to some extent. However, lightweight monitoring measures, such as web behavior tracking or interaction logging, may still be valuable for understanding the extent and nature of participants' reliance on LLMs. Rather than serving as a basis for exclusion, such information could contextualize the collected data and support more informed interpretation of the results.

\paragraph{Post-collection considerations.}
Post-collection considerations in such settings should therefore focus less on exclusion and more on understanding how varying 
levels of LLM assistance may influence the collected data. Disclosed or inferred LLM usage may then be incorporated as a contextual
variable during analysis, for example by comparing patterns across responses with differing levels of reported assistance and
evaluating whether study findings remain consistent across these conditions. Transparent reporting of how such information was 
documented and incorporated into the analysis is therefore important.

\subsubsection{Unrestricted LLM Usage} 
Not all crowdsourcing studies require human-only responses. Unrestricted LLM usage is warranted when the study explicitly requires workers to interact with or prompt LLMs, for instance to examine prompting strategies, generation patterns, or human-AI collaboration. In such contexts, LLM use is not a confound but a core part of the research design. This was the second-largest position in our survey ($33.3\%, n=40$), with respondents identifying studies of LLM use itself ($26.7\%, n=32$) and human-AI interaction research ($7.5\%, n=9$) as primary contexts where LLM use is appropriate.

\section{Conclusion \& Future Work}

 Given the increased adoption of LLMs, crowdworkers hired for data creation may be inclined to outsource text generation tasks to a model, either to `game the system' or because it has become part of their writing practice.
Given the longstanding assumption that crowdsourcing collects human data, this raises the question: Can crowdsourcing survive the LLM era?
Despite the pressing nature of the question, we were lacking a shared understanding of the research community's perspective on this. Therefore, we conduct a survey, collecting a total of $155$ responses from researchers who use crowdsourcing in NLP and related fields. Subsequently, we analyze their experiences and opinions collecting free-text responses via crowdsourcing in the LLM era~(Secs.~\ref{sec: RQ1}, \ref{sec: RQ2}).

 The results indicate that LLM-usage by crowdworkers has be taken as a realistic possibility: $44.0\%$ of respondents who crowdsourced free-text responses reported observing LLM usage in their data.
Our survey respondents have nuanced perspectives on the problem. While respondents raised concerns about result reliability, validity and homogenization, many emphasize that crowdsourcing as a method can remain useful depending on the task type (e.g., short text production tasks, crowdworkers in a verifier role) and incentive structure (adequate compensation, meaningful tasks). Overall, the results indicate that crowdsourcing can only `survive' in the LLM era if we no longer use it as quick data collection or validation method. Instead, its applicability has to be carefully considered, and study setup and data analyses have to be mindfully designed. Our respondents emphasize that for transparency and reproducibility, any action or precaution taken in this regard should be well documented.
In addition to our analytical results about challenges and opinions, the paper therefore provides a set of practical considerations that we derive from the survey insights~(Sec.~\ref{rq3}). They aim to guide future research designing and documenting crowdsourcing studies in the LLM era.

 Our findings point to several open questions that extend beyond researchers' perceptions. While concerns about LLM-assisted crowd work are widespread, we still know little about how and when crowdworkers use LLMs in practice, or how this behavior varies across platforms and task types. Future work should therefore focus on building a more empirical understanding of AI-assisted crowd work. This includes the different ways workers incorporate LLMs into their workflows, the resulting impact on data quality and the extent to which existing microwork practices, incentives, and working conditions encourage or even necessitate the use of LLMs. We also see value in broadening the conversation to include the perspectives of crowdworkers and platform providers, whose experiences are essential for understanding how crowdsourcing is evolving in the LLM era. This could help further extend the considerations discussed in the study to serve as the foundation for community-driven best practices on the use of crowdsourcing in the age of LLMs, helping researchers make informed decisions about study design, data collection, and reporting. In the longer term, these efforts could contribute to broader community recommendations for responsible human data collection.

\section*{Limitations}

Our survey received responses from 155 researchers. While this does not permit statistically representative claims about the entire NLP community, the sample size is comparable to the 135--327 respondents reported in recent NLP community surveys \citep{michael-etal-2023-nlp,mosbach-etal-2024-insights,thomson-etal-2025-evolving, hojer-etal-2025-research, romberg-etal-2026-reassessing}. Accordingly, our findings should be interpreted as descriptive of the perspectives represented in our sample rather than definitive estimates of community-wide attitudes.

Since we adopted an adhoc recruitment process and participation was voluntary, 
researchers who perceive LLM usage by crowdworkers as a problem were more likely
to participate, potentially skewing responses toward greater concern
and awareness than exists in the broader community. Furthermore,
engineering sciences, which subsumes NLP under the DFG classification,
is the dominant respondent category, and the *ACL-based recruitment
further accentuates this skew, meaning the views expressed are more
representative of the NLP community than of adjacent fields such as
psychology or the social sciences. Finally, given the rapid pace of
LLM development and the extended data collection period, some findings
may already have shifted and are likely to continue evolving.

\section*{Acknowledgments}
We are extremely grateful to everyone who took the time to participate in and distribute our survey, thereby enabling this research. Thanks to your insights we now have a more nuanced understanding of the current state and the future of crowdsourcing. 

This work was supported by the Belgian American Educational Foundation (BAEF) and the Research Foundation Flanders (FWO-Vlaanderen) under grant number 1S96324N, the SoftwareCampus project
Placebo, funded by German Federal Ministry of Research, Technology and Space (BMFTR) under the grant 01IS23072, and by  Villum Fonden, research grant ([VIL60860]).

\section*{Ethics Statement}
This study involved collecting opinions and experiences from
researchers via an online survey. Participation was voluntary,
anonymous, and involved no sensitive personal data beyond
professional background, field of expertise, and demographics (age, gender, country of employment). 
The study does not involve vulnerable
populations, sensitive topics, or data that could cause harm to
participants. Informed consent was obtained from all participants
prior to participation, including agreement to voluntary
participation and data privacy provisions. Email contact information was sourced solely from publicly available
academic publications.

\section*{AI Usage Statement}

We used AI assistance during the preparation of this manuscript for text editing, literature search, and generating code for data analysis and visualization. All AI-generated outputs were critically reviewed, verified, and revised by the authors.

\bibliography{custom}
\clearpage
\onecolumn
\appendix

\section{Appendix}

\subsection{Survey methodology}
\subsubsection{Identifying and Contacting Authors}\label{append:contact authors}

As mentioned in Section~\ref{sec:survey_method}, to recruit survey participants, we also contacted researchers who had recently published work involving human data annotation, crowdsourcing, or free-text data collection at major computational linguistics venues. We describe below how we collected the email lists to reach out to potential participants for this survey: 

\paragraph{Paper Collection.} We collected metadata (titles, authors, and paper URLs) and full-text PDFs for all papers published in $2024$ and $2025$ across 6 *ACL-affiliated venues: ACL, NAACL, EMNLP, EACL, AACL-IJCNLP, and TACL, including both main conference and Findings tracks from the ACL Anthology\footnote{\url{https://aclanthology.org}}.  In total, this yielded approximately $12.6k$ successfully retrieved papers across all venues.

\paragraph{Relevance Filtering.}
We employed three complementary strategies to identify papers likely to involve human annotation or crowdsourced data collection, and took their union (deduplicated by paper ID and title) as the final candidate set.

\textit{Keyword-based full-text search.}
We extracted text from each PDF using PyMuPDF,\footnote{\url{https://pymupdf.readthedocs.io}}
excluded the references section via regex-based boundary detection, and searched for a curated list of keywords. These included specific crowdsourcing platform names (e.g., ``Amazon Mechanical Turk'', ``Prolific'') and general annotation-related terms (e.g.,
``crowdsourcing'', ``human annotators'').

\textit{LLM-based classification.}
We used Meta's Llama~3.1 8B Instruct model~\cite{grattafiori2024llama} to classify papers based on their content. The model was prompted with the paper text and asked: 

\begin{quote}

\textit{From the abstract [and introduction], does this paper involve any human data annotation, crowdsourcing, human data collection, free-text human responses, or discussion of annotation quality?}

\textit{Respond with ONLY one word: YES or NO.}
\end{quote}

We ran this classification twice: once using only the abstract, and once using both the abstract and introduction (extracted from the first three pages of each PDF via regex-based section boundary detection), to capture papers where annotation details appear only in the introduction. Both resulting candidate lists were included in the final union. Inference was performed locally via Ollama\footnote{\url{https://ollama.com/}} with default parameters on an Apple M4 Pro ($48$\,GB), totaling   ${\sim}11$ hours.

\textit{EMNLP $2025$ Resource Track.}
For EMNLP $2025$, we additionally had access to the publicly available conference program, which included track-level metadata. We filtered for papers in the ``Resources and Evaluation'' track, yielding $365$ papers which were inherently likely to involve dataset construction and human annotation as compared to other papers. This track-level filtering was uniquely available for EMNLP $2025$; no other venue in
our collection published comparable program metadata at the time of data collection.

\paragraph{Contact Extraction.} Author names were obtained from the BibTeX metadata on the ACL Anthology. Email addresses were extracted from the first two pages of each PDF using regular expressions, handling both standalone addresses and grouped patterns
(e.g., \texttt{\{a,b\}@domain}). After relevance filtering, 2,276 papers were identified as candidates. We contacted the authors of all papers for which at least one valid email address could be extracted from the paper PDF.

\clearpage
\subsubsection{Survey Workflow}\label{append:survey-flow}

\vspace{6pt}

\setlength{\tabcolsep}{6pt}
\renewcommand{\arraystretch}{1.35}
\rowcolors{2}{rowScreen}{white}
\begin{longtable}{>{\bfseries}p{1cm} p{0.55cm} p{8cm} p{4cm}}
\toprule
\textbf{Section} & \textbf{ID} & \textbf{Question} & \textbf{Routing / Condition} \\
\endfirsthead
\toprule
\textbf{Section} & \textbf{ID} & \textbf{Question (summary)} & \textbf{Routing / Condition} \\
\endhead

\sechead{secExp}{Screening}
 & Q1 & Have you ever conducted a crowdsourcing study? &
  \cond{If No $\rightarrow$ skip to Opinions} \\
 & Q2 & Did participants generate free-text responses? &
  \cond{If No $\rightarrow$ skip to Opinions} \\

\sechead{secExp}{Experience \quad {\normalfont\small(shown only when Q1 \& Q2 = Yes)}}
 & E1 & Years of experience in annotation studies? & \\
 & E2 & Years of experience with crowdsourcing? & \\
 & E3 & Which crowdsourcing platforms have you used? & \\
 & E4 & In which languages have you conducted studies? & \\
 & E5 & Primary purpose of your crowdsourcing studies? & \\
 & E6 & Did you analyse the free-text responses? &
  \cond{If No $\rightarrow$ skip E7} \\
 & E7 & Type of analysis performed \& objectives? &
  \cond{Shown only if E6 = Yes} \\
 & E8 & Have you observed crowdworkers using LLMs? & \\
 & E9 & What signs suggested LLM use? &
  \cond{If answer includes ``used detection tools'' $\rightarrow$ show E10} \\
 & E10 & Which detection models/tools did you use, and how reliable? &
  \cond{Shown only if E9 includes detection tools} \\
 & E11 & Actions taken regarding the suspected worker? & \\
 & E12 & What was done with suspected LLM-generated data? & \\
 & E13 & Which task types were most affected by LLM use? & \\
 & E14 & Platform(s) with most LLM-generated content? & \\
 & E15 & Platform(s) with least LLM-generated content? & \\
 & E16 & Did you anticipate crowdworkers using LLMs? &
  \cond{No $\rightarrow$ E21 \quad Yes + precautions $\rightarrow$ E18 \quad Yes, no precautions $\rightarrow$ E17} \\
 & E17 & Why did you decide not to adapt the study? &
  \cond{Shown only if E16 = ``Yes, no precautions''} \\
 & E18 & Which precautions did you take? &
  \cond{Shown only if E16 = ``Yes + precautions''. If includes warning $\rightarrow$ E19, else $\rightarrow$ E21} \\
 & E19 & Do you think adding a warning was a good idea? &
  \cond{Shown if E18 includes warning. If No $\rightarrow$ E20, else $\rightarrow$ E21} \\
 & E20 & Why do you think the warning was not a good idea? &
  \cond{Shown only if E19 = No} \\
 & E21 & Has the prevalence of LLM-generated content led you to change your approach? & \\

\sechead{secExp}{Opinions}
 & O1 & How does LLM-generated text affect research quality \& validity? & \\
 & O2 & Ways to discourage or detect LLM use in responses? & \\
 & O3 & Should platforms allow, restrict, or monitor LLM use? & \\
 & O4 & What platform-level support would help manage LLM use? & \\
 & O5 & Situations where LLM use by crowdworkers is acceptable? & \\
 & O6 & Additional thoughts on LLM use by crowdworkers? & \\

\sechead{secExp}{Respondent Info}
 & D1 & Gender & \\
 & D2 & Age & \\
 & D3 & Country of current employment & \\
 & D4 & Current role & \\
 & D5 & Scientific discipline & \\
 & D6 & Main research topics or focuses & \\

\sechead{secExp}{Other}
 & T1 & Did you use GenAI to fill (parts of) this survey? & \\
 & T2 & Feedback or comments about this study? & \\

\bottomrule
\end{longtable}

\clearpage

\subsubsection{Participant Consent} 
\begin{figure}[h]
    \centering
    \includegraphics[width=.85\linewidth]{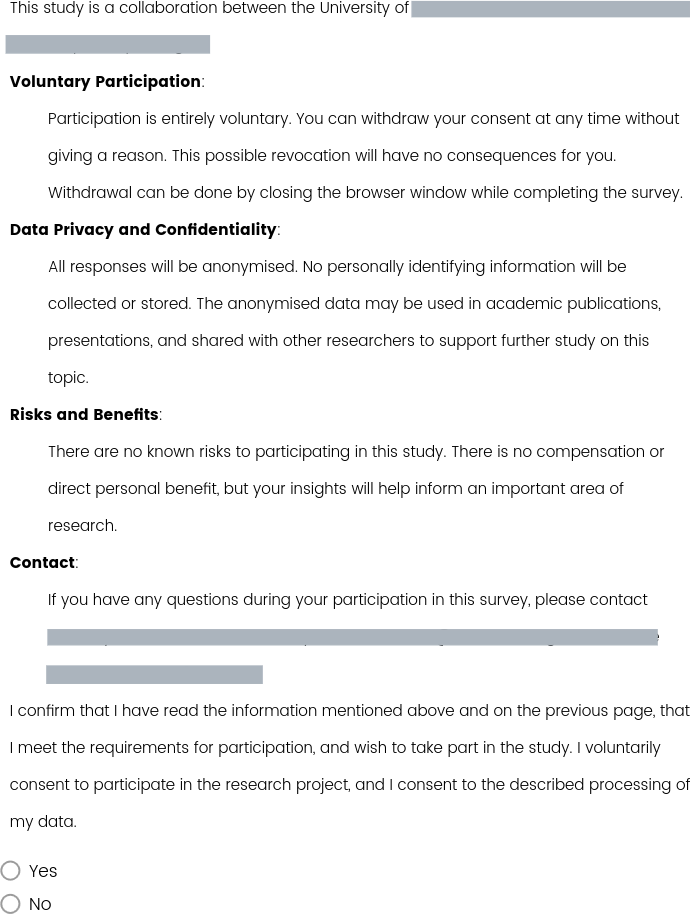}
    \caption{Consent page of survey}
    \label{fig:consent}
\end{figure}

As shown in Fig. \ref{fig:consent}, informed consent was obtained from
all participants prior to taking the survey, including the option
to withdraw at any time without consequence.

\clearpage

\clearpage

\subsection{RQ2 additional material: Perceived impact of LLM-generated responses} \label{appendix-rq2}
Table~\ref{tab:themes-impact} lists all themes identified in the responses to the survey question `How do you think the increasing presence of LLM-generated text in crowdsourced datasets affects research that relies on human-written data?'. We provide representative quotes from the survey participants to illustrate the themes. 

\begin{figure*}[!hb]
    \centering
    \includegraphics[width=0.98\linewidth]{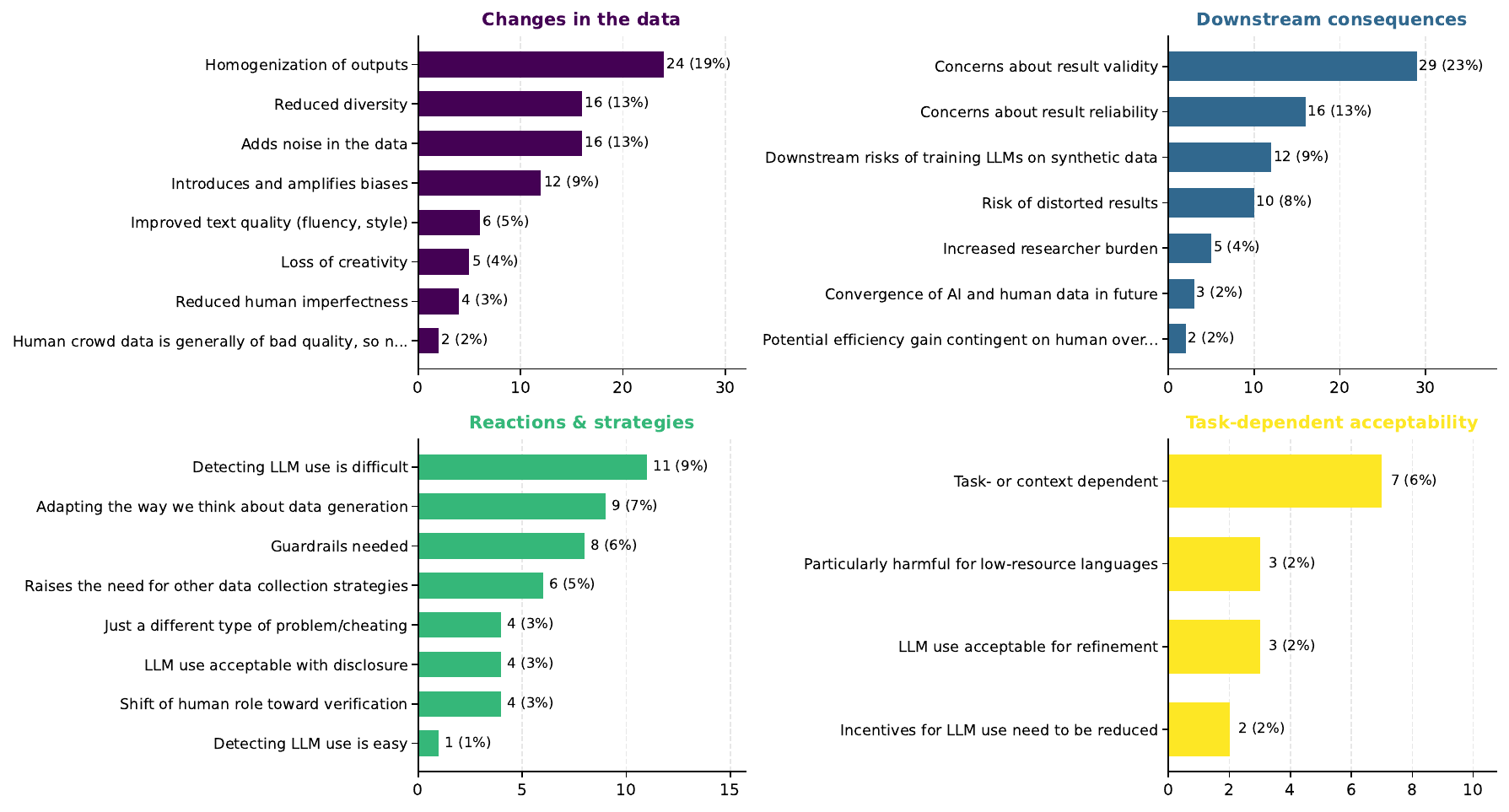}
    \caption{Perceived impact themes grouped by category ($n = 127$, multi-label).}
    \label{fig:impact_themes}

\end{figure*}

In addition, Figure~\ref{fig:impact_themes} shows the frequency of each theme across survey responses.

\newcommand{\axisrow}[1]{%
  \midrule
  \multicolumn{2}{@{}l}{\textit{#1}} \\
  \cmidrule(lr){1-2}}

\footnotesize\footnotesize
\begin{longtable}{p{4.2cm}p{0.68\textwidth}}
\toprule
\textbf{Theme} & \textbf{Example quote} \\

\axisrow{Changes in the data}

 Homogenization of outputs &
   ``Using LLMs may also homogenize views and textual responses'' \\
 Introduces and amplifies biases &
   ``it reinforces the LLM biases and circles the bias back into research instead of introducing diversity and novelty'' \\
 Improved text quality (fluency, style) &
   ``The language fluency has improved, but the content may be influenced by the large model.'' \\
 Adds noise in the data &
   ``It adds a lot of slop.'' \\
 Reduced diversity &
   ``I also worry that it will further skew data as to not reflect minority groups [\ldots] which in turn actively represses their identities.'' \\
 Reduced human imperfectness &
   ``Data quality may deviate from raw unpolished version human-generated text.'' \\
 Loss of creativity &
   ``I believe it will hurt the emotional depth, creativity and honesty, making the data less representative of the human experience.'' \\

Human crowd data is generally of bad quality, so not much is lost &
   ``Crowd-sourced annotation already represented a prioritization of cost and scale over quality or validity''
   \\

\axisrow{Downstream consequences}

 Concerns about result reliability &
   ``the fact that we can never be absolutely certain of the quality of the work significantly impacts the reliability of the data obtained''
   \\
 Concerns about result validity &
   ``LM responses may not represent the true perspective of participants, and so the insights drawn from research would be an incorrect representation of the population.''
   \\
 Risk of distorted results &
   ``potentially distorting study outcomes and effect estimates'' \\
 Downstream risks of training LLMs on synthetic data &
   ``we might end up with aligning to a wrong target'' \\
 Increased researcher burden &
   ``we need to invest more effort to ensure heterogeneity (thereby get the real value),'' \\
 Convergence of AI and human data in future &
   ``perhaps the generic human writing style will eventually become more AI-styled because many will use AI to write. At the same time, AI will become more personalized and will adopt human styles as well.'' \\

Potential efficiency gain contingent on human oversight &
   ``LLMs can improve efficiency, but their impact on data quality and validity depends on how they are integrated and whether human judgment remains central to the process'' \\

\axisrow{LLM-generated content: Reactions \& strategies}

 Raises the need for other data collection strategies &
   ``To mitigate these issues, I prefer working with a smaller in-house annotation group [\ldots]'' \\
 Guardrails needed &
   ``careful participant tracking and validation are important'' \\
 Adapting the way we think about data generation &
   ``the rationale for collecting `purely human' is increasingly difficult to defend and enforce, as humans---especially untrained crowdsourced humans are increasingly worse than LLMs at performing many of these tasks.'' \\
 Shift of human role toward verification &
   ``it might ease cognitive load of workers, allowing for more or more thorough checking if facilitated well.'' \\
 LLM use acceptable with disclosure &
   ``I viewed LLM as a boon, but over the period of time, their overuse is leading to lack of authenticity when not disclosed.'' \\
 Detecting LLM use is easy &
   ``[\ldots] so if people use LLMs to do our task, they'll just submit nonsense [\ldots]. It's annoying but easy to filter out.'' \\
 Detecting LLM use is difficult &
   ``it has become almost impossible to determine whether a study has been completed by a human participant, machine or combination thereof''
   \\
 Just a different type of problem/cheating &
 ``Same as before, people cheated all along.'' 
   \\

\axisrow{(Acceptability of) LLM use is task dependent}

 LLM use acceptable for refinement &
   ``If people use LLMs to just improve clarity and fix grammar, I don't mind"
   \\
 Task- or context dependent &
   ``So for rare and interesting tasks keep it, for repetitive ones checking human performance may need to monitored [sic: monitor]'' \\
 Incentives for LLM use need to be reduced &
   ``Tasks that feel meaningful to the annotator and are ideally also adequately paid tend to lead to better results.'' \\
 English data is most affected &
   ``for non-English languages, for underrepresented languages, the problem is still not so huge.'' \\
 Particularly harmful for low-resource languages &
   ``concerned about the integrity of low-resource language datasets, where `cheated' or machine-generated data is easier to submit and harder to detect.''
   \\
\bottomrule
\caption{Themes extracted from the researchers' responses to the question `How do you think the increasing presence of LLM-generated text in crowd-sourced datasets affects research that relies on human-written data?'. We provide example quotes from the survey respondents.}
\label{tab:themes-impact}
\end{longtable}

\clearpage
\newpage

\subsubsection{Theme extraction prompt}
\label{app:prompts}

The following prompt was used with \texttt{qwen3.5\_27b} with default parameters to independently extract themes related to the perceived impacts of LLM-generated text in crowdsourced datasets from each survey response in an open-ended manner, without a predefined theme list. This was used during the initial theme identification stage described in Section~\ref{sec: RQ2}. Inference was performed locally via Ollama on an Apple M4 Pro ($48$ GB), averaging $15$ seconds per response (${\sim}30-40$ minutes total).

\begin{center}
\begin{tcolorbox}[colback=gray!20, colframe=gray!60, boxrule=0.8pt, width=0.9\columnwidth, boxsep=3pt, title=Model Prompt - Theme Extraction for LLM Impacts, breakable]
You are an expert in social and computational linguistics specializing in thematic analysis of survey responses.

\medskip
\#\# Task

Analyze a free-text response to the following survey question and assign one or more themes to each response.

\medskip
**Survey question:**

"How do you think the increasing presence of LLM-generated text in crowdsourced datasets affects research that relies on human-written data? (e.g., data quality, validity, impacts on study outcomes)"

\medskip
**Survey context:**

This question was part of a research study investigating whether and to what extent crowd workers use LLMs to answer free-text surveys, and how researchers experience and respond to this phenomenon.

\medskip
\#\# Instructions

1. Read the response carefully.

2. Identify ALL themes expressed in the response — a response may express multiple themes.

3. For each assigned theme, quote the phrase(s) or sentence(s) from the response text that expresses the theme.

\medskip
\#\# Output Format

Return the results in valid JSON with no explanation, markdown, or extra text. Use the following structure:

{"themes": [{"theme": "<theme name>", "quote": "<quoted text>"}]}
    
\end{tcolorbox}
\end{center}

\clearpage

\subsection{RQ3 additional material: Examples for acceptable LLM use}
\label{app:acceptable_quotes}

Table~\ref{tab:acceptable_themes} lists the themes identified from responses to question O5, grouped by acceptability category, with representative quotes.

\begin{table*}[t]
\centering
\small
\begin{tabular}{p{0.18\linewidth} p{0.04\linewidth} p{0.7\linewidth}}
\toprule
\textbf{Theme} & \textbf{\#} & \textbf{Example quote} \\
\midrule
\multicolumn{3}{l}{\textit{Restrict Use --- Human signal is essential}} \\
\midrule
Not acceptable & 14 & ``If it's to be done by LLM the researcher can just use the LLM directly. It's not crowdsourcing." \\
\midrule
\multicolumn{3}{l}{\textit{Controlled Use --- LLM assistance may be partially acceptable}} \\
\midrule
Editing & 22 & ``Polishing (in terms of language) user answers that contain their private opinions." \\
Miscellaneous tasks & 11 & ``Maybe for tasks that require searching the web, as a searching tool (e.g. deep research)." \\
Factual tasks & 10 & ``Classification of objects, or any task where there are certain answers that are true could be tasks where LLMs can be used with less problems since the answer is not subjective." \\
Human oversight & 10 & ``LLMs can provide preliminary annotations for various tasks. Humans can later be used to curate and correct the LLM annotation." \\
Accessibility support & 6 & ``Users with linguistic handicaps that use LLMs to correct/reformulate for clarity their writing (dyslectics)." \\
Data augmentation & 5 & ``In data scarcity study domains it's necessary to use LLM." \\
Everyday workflows & 4 & ``Researchers might want to look at how people, in 2026, complete certain tasks -- LLMs are increasingly a part of day-to-day life, so it might be acceptable." \\
\midrule
\multicolumn{3}{l}{\textit{Unrestricted Use --- LLM interaction is itself the study object}} \\
\midrule
Study LLM use & 32 & ``If you wanted to study how people used some LLM (like a user-study), this could make sense." \\
Human-AI interaction & 9 & ``Only for studies evaluating human-LLM interaction." \\
\midrule
\multicolumn{3}{l}{\textit{Ambiguous --- No clear position}} \\
\midrule
Unsure & 7 & ``I would think so but I can't come up with a specific example." \\
Depends on study goals & 5 & ``It really depends on the researcher's hypothesis and the centrality of real human responses to testing that hypothesis." \\
\bottomrule
\end{tabular}
\caption{Themes extracted from researchers' responses to the question ``Are there specific tasks, types of responses or situations in the context of crowdsourcing studies where the use of large language models (LLMs) could be considered acceptable?". Themes are grouped by acceptability category. Counts (\#) reflect the number of responses mentioning each theme out of $n=120$ substantive responses (multi-label coding).}
\label{tab:acceptable_themes}
\end{table*}

\end{document}